\documentclass[conference]{IEEEtran}
\IEEEoverridecommandlockouts
\usepackage[small]{caption}
\usepackage[utf8]{inputenc}
\usepackage{amssymb,amsmath,amsthm,amsfonts}
\usepackage{comment}
\usepackage{multirow}
\usepackage{mathtools}
\usepackage[linesnumbered,ruled,vlined]{algorithm2e}
\usepackage{graphicx}
\def\BibTeX{{\rm B\kern-.05em{\sc i\kern-.025em b}\kern-.08em
    T\kern-.1667em\lower.7ex\hbox{E}\kern-.125emX}}
\usepackage{footnote}    
\usepackage{color}   
\usepackage{booktabs}
\DeclareMathOperator*{\argmax}{argmax}

\newcommand*{\jy}[1]{\textcolor{blue}{#1}}
\newtheorem{lemma}{Lemma}
\newtheorem{definition}{Definition}
\begin{document}

\title{\textsc{ErGAN}: Generative Adversarial Networks for Entity Resolution\footnote{Shot paper published by ICDM 2020}
}

\author{\IEEEauthorblockN{Jingyu Shao\IEEEauthorrefmark{1}, Qing Wang\IEEEauthorrefmark{1}, Asiri Wijesinghe\IEEEauthorrefmark{1}, Erhard Rahm\IEEEauthorrefmark{2}}
    \IEEEauthorblockA{\IEEEauthorrefmark{1}Research School of Computer Science, the Australian National University
    \\\{Jingyu.Shao, Qing.Wang and Asiri.Wijesinghe\}@anu.edu.au}
    \IEEEauthorblockA{\IEEEauthorrefmark{2}Database Group, University of Leipzig
    \\Rahm@informatik.uni-leipzig.de}\thanks{This paper is published by ICDM 2020}
}
\vspace{1cm}

\maketitle

\begin{abstract}
  Entity resolution targets at identifying records that represent the same real-world entity from one or more datasets. A major challenge in learning-based entity resolution is how to reduce the label cost for training. Due to the quadratic nature of record pair comparison, labeling is a costly task that requires a significant effort from human experts. However, without sufficient training data, a powerful machine learning model may be overfitting. This challenge is further aggravated when the underlying data distribution is highly imbalanced, which commonly occurs in entity resolution applications. Inspired by recent advances of generative adversarial network (GAN), in this paper, we propose a novel deep learning method, called \textsc{ErGAN}, to address the challenge. \textsc{ErGAN} consists of two key components: a label generator and a discriminator which are optimized alternatively through adversarial learning. To alleviate the issues of overfitting and highly imbalanced distribution, we design two novel modules for diversity and propagation, which can greatly improve the model generalization power. We theoretically prove that \textsc{ErGAN} can overcome the model   collapse and convergence problems in the original GAN. We also conduct extensive experiments to empirically verify the labeling and learning efficiency of \textsc{ErGAN}. The experimental results show that \textsc{ErGAN} beats all state-of-the-art baselines, including unsupervised, semi-supervised, and unsupervised learning methods.
  \end{abstract}


\section{Introduction}

Entity Resolution (ER) is an important and ubiquitous component of real-world applications in various fields, such as national census, health sector, crime and fraud detection, bibliographic statistics, and online shopping \cite{christen2012data}. For example, if a national census agency wants to obtain the population growth of its country in a time period, ER is necessary to detect whether records from different time points refer to the same person, no matter whether he or she changed name (e.g. due to marriage), postal address and so on.
Learning-based ER methods have been widely used in the past years.
However, due to the quadratic nature of record pair comparison required by ER tasks \cite{christen2008automatic}, labeling is costly, time consuming, and highly imbalanced. 
This raises the difficulty of applying supervised learning methods for ER because supervised learning methods require a sufficient number of labeled instances for training, which is infeasible in many real-world applications. To reduce the labeling effort, a number of semi-supervised learning methods have been proposed \cite{kejriwal2015semi, wang2015efficient, shao2019learning}. 

Generally, semi-supervised learning aims to train a machine learning model with both labeled instances (called seeds) and unlabeled instances \cite{zhu2005semi}.
Several semi-supervised learning methods have been proposed based on a low-density separation assumption, i.e. there exists a low-density ``boundary'' so that instances belonging to different classes can be distinguished \cite{basu2002semi, li2017semi}. 
However, such a boundary may not always exist or can be clearly identified, especially when the number of labeled instances is small  \cite{jain2010data}.
Some semi-supervised learning methods have ultilized the idea of self-learning, which firstly trains a classifier using labeled instances, and then selects unlabeled instances with predicted labels to train a classifier iteratively \cite{kejriwal2015semi}. Although promising, these methods often lead to the issue of overfitting when labeled instances in training are limited \cite{pyle1999data}.


\medskip
In this paper, we focus on tackling the following two challenges, which cannot be handled by the existing ER methods: (1) \textbf{the overfitting problem}; (2) \textbf{the imbalanced class problem}. The overfitting problem happens when the number of labeled instances is limited and a learning model is powerful enough to remember all the features of training instances. In such cases, the learning model can correctly predict the classes of seen instances with high certainty, but fail to predict the classes of unseen instances, thus losing the generalization ability. For the imbalanced class problem, it is due to the fact that the number of matches (record pairs referring to the same entity) is far less than the number of non-matches in ER tasks. For example, given two datasets with the number of records $n$ and $m$, respectively, the largest number of matches between them is $\texttt{min}(n,m)$ while the largest number of non-matches is $n*m$. 
Traditionally, blocking techniques can help to alleviate the imbalanced class problem by grouping potentially matched instances into the same cluster. However, selecting a blocking method also requires prior knowledge or sufficient training instances \cite{wang2016semantic,shao2018active}, which is still hard to achieve under a very limited number of training instances. 

\begin{figure*}[ht!] 
		\centering
	\begin{center}
		\includegraphics[width =1 \textwidth]{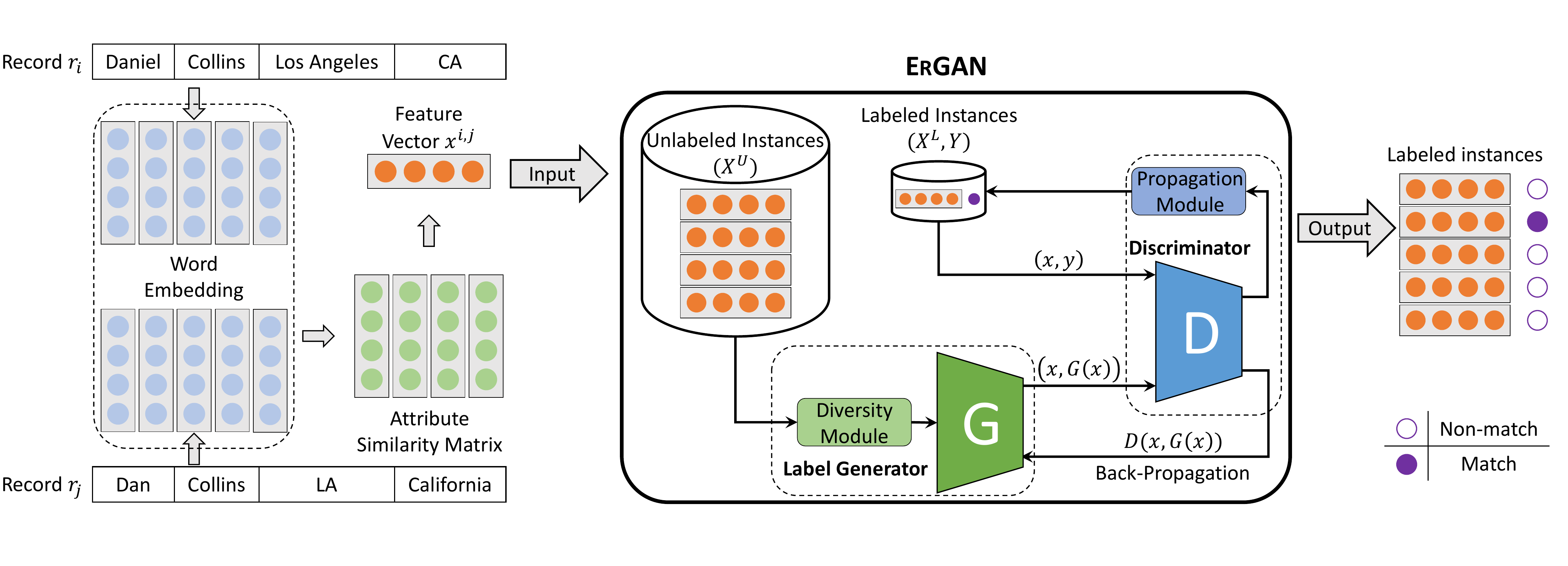}
	\end{center}
	\vspace{-6mm}
	\caption{\textbf{Overview of our framework.} Given a record pair $r_i$ and $r_j$, each record is first represented as a matrix of vector embeddings using word embedding. Based on this, the record pair is transformed into an attribute similarity matrix, which leads to a feature vector (i.e., an unlabeled instance). Using only a limited number of labeled instances for training, \textsc{ErGAN} takes unlabeled instances as input and classifies them as being matches or non-matches (i.e. predicting their labels).} 
	\label{fig_sampling1}\vspace{-3mm}
\end{figure*}

\medskip

Generative adversarial network (GAN) and its variants have recently emerged as a powerful deep learning technique for real-world applications across various domains such as image generation and natural language processing \cite{goodfellow2014generative,goodfellow2016deep}. 
Inspired by these advances, in this paper, we develop a novel generative adversarial network, called \textsc{ErGAN}, to solve the aforementioned challenges faced by ER applications. In \textsc{ErGAN}, there are two key components: (1) a \emph{label generator} $G$ that aims to generates pseudo labels for unlabeled instances, and (2) a \emph{discriminator} $D$ that aims to distinguish instances with pseudo labels from instances with real labels. 
The discriminator $D$ is trained using not only a small number of instances with real labels but also a large number of instances with high-quality pseudo labels. However, the question arises: {how to ensure the high-quality of pseudo labels generated for unlabeled instances?} Unfortunately, the existing GAN and its variants cannot guarantee this when the number of instances with real labels is limited. To address this question, our model \textsc{ErGAN} is designed to incorporate two modules: \emph{diversity module} and \emph{propagation module} into the label generator $G$ and the discriminator $D$, respectively. The diversity module enriches the diversity of unlabeled instances during the sampling process, while the propagation module guarantees that only unlabeled instances with high-quality pseudo labels can be propagated into the training of the discriminator $D$. We formally prove that $G$ and $D$ converge to the equilibrium point, achieving the global optimality. 
Figure \ref{fig_sampling1} shows an overview of our framework of using \textsc{ErGAN} for ER tasks.

%

\textsc{ErGAN} can alleviate the overfitting problem. This is because, instead of training a single classifier, \textsc{ErGAN} trains the label generator $G$ and the discriminator $D$ adversarially such that $D$ improves $G$ to generate pseudo labels for unlabeled instances and $G$ regularizes $D$ from overfitting through propagating unlabeled instances with high-quality pseudo labels. In essence, $G$ serves as an adaptive regularization term acting on $D$ during the minibatch training process. 
\textsc{ErGAN} can also alleviate the imbalanced class problem. We observe that, the key to solving the imbalanced class problem lies in how to effectively approximate the true distribution using limited instances with real labels. Driven by this observation, \textsc{ErGAN} employs the diversity module to sample instances diversely from different feature subspaces.
This leads to a dual boosting: both improving label efficiency of selecting instances from the minority class, and improving learning efficiency of learning a fast approximation on the underlying data distribution.

\medskip
\noindent\textbf{Contributions.} 
Our contributions are summarized as follows:

\begin{itemize}
    \item \textbf{We propose a semi-supervised generative adversarial network, namely \textsc{ErGAN}, for entity resolution.} \textsc{ErGAN} has two key components: a label generator and a discriminator, which are optimized in an adversarial learning manner. As \textsc{ErGAN} is specifically designed for ER classification, it can be used jointly with any word embedding or string matching techniques for ER tasks.
    \item \textbf{We develop two integrated modules: diversity and propagation modules to improve the model generalization ability.} Consequently, even when only a very limited number of labeled instances are available, \textsc{ErGAN} can still effectively infer the true distribution of all labels in a semi-supervised manner. 
    \item \textbf{We theoretically prove that \textsc{ErGAN} overcomes the model collapse and convergence problems in the original GAN.} Specifically, the following properties of \textsc{ErGAN} are proven: (1) The global optimality can be guaranteed; (2) The label generator $G$ and the discriminator $D$ converge to the equilibrium point; (3) The diversity module helps improve the generalization without compromising the global optimality and equilibrium; and (4) The mode collapse issue is alleviated in \textsc{ErGAN}.
    \item \textbf{We conduct extensive experiments to empirically verify the effectiveness of \textsc{ErGAN}.} The experimental results show that \textsc{ErGAN} outperforms all state-of-the-art baselines, including unsupervised, semi-supervised and full-supervised learning methods. Even when the label cost is very limited and the state-of-the-art baselines fail to predicate labels, \textsc{ErGAN} can still achieve reasonably good performance.
\end{itemize}
It is worthy to note that, although we only consider the use of \textsc{ErGAN} for entity resolution in this paper, the techniques of \textsc{ErGAN} for handing overfitting and imbalanced data can be much more widely applicable.

\smallskip
\section{Problem Formulation}
Let $R$ be an ER dataset consisting of a set of records where each $r\in R$ is associated with a number of attributes $A$. We use $r.a$ to refer to the value of an attribute $a\in A$ in a record $r$.  Each record pair $(r_i, r_j)$ in $R$ corresponds to a feature vector $x^{(ij)}\in \mathbb{R}^m$ where each dimension $x^{(ij)}_k$ indicates a feature value, e.g., the textual similarity of values in an attribute $a_k\in A$, i.e., $x^{(ij)}_k=f(r_i.a_k, r_j.a_k)$, calculated by a function $f$. 

Let $X=\{x^{(ij)}|(r_i,r_j)\subseteq R\times R\}$ be the set of all instances corresponding to record pairs in $R$ and $Y=\{M,N\}$ be a label space where $M$ and $N$ refer to two labels \emph{match} and \emph{non-match}, respectively.  There is a small subset $X^L\subseteq X$ of instances that are labeled, while the other instances in $X$ are unlabeled, i.e., $X^U=X-X^L$. We assume $|X^L| <\!\!< |X^U|$, i.e., $X$ has a very limited number of labeled instances in $X^L$ but a large number of unlabeled instances in $X^U$. We denote $(X^L, Y)$ as a set of instances in $X^L$ and their labels in $Y$, and  $(x^L,y)\sim (X^L, Y)$ as a pair of  instance $x^L\in X^L$ and its label $y\in Y$. Our task is to tackle the ER classification problem as formulated below.
\begin{definition}
 Given a set $X$ of instances with $X=X^L\cup X^U$ and $|X^L| <\!\!< |X^U|$, and a label space $Y=\{M,N\}$, the \textbf{ER classification problem} is to learn a model $\Lambda$ that can predict a label $\hat{y}\in Y$ for each unlabeled instance $x\in X^U$ w.r.t. 
 \begin{equation}
     \text{max  } E(\Lambda)/|X^U|
 \end{equation}where $E(\Lambda)=\sum\limits_{x\in X^U\wedge \hat{y}=y} 1$.
 \end{definition}
 
Intuitively, $E(\Lambda)$ refers to the total number of unlabeled instances in $X^U$ whose labels are correctly classified by $\Lambda$. 
\smallskip

\section{Proposed Method: \textsc{ErGAN}}

Our proposed method \textsc{ErGAN} consists of two components: (1) a \emph{label generator} $G$; and (2) a \emph{discriminator} $D$. Both $G$ and $D$ are differentiable functions. 

\subsection{Label Generator}

In \textsc{ErGAN}, a label generator $G$ can obtain instances from $p(X^U)$, but does not know about $p(Y)$ nor $p(X,Y)$. Nevertheless, we know that $p(X^U)\approx p(X)$ because $X^U\subseteq X$ and $|X^U|/|X|$ is close to 1.
The goal of $G$ is to learn a conditional distribution $p_g(Y|X^U)\approx p(Y|X^U)$, i.e., given an instance $x\sim p(X^U)$ as input, $G$ generates a pseudo label $\hat{y}$ for $x$. Ideally, the pseudo label $\hat{y}$ generated for an instance $x$ by $G$ should be the same as the real label of $x$.
To simulate the conditional distribution $p(Y|X^U)$, the label generator $G$ receives feedback (i.e. gradients) from the discriminator $D$ and is trained iteratively through backpropagation.  

\medskip\noindent\textbf{Diversity module.~} One major difference of our \textsc{ErGAN} from the original GAN and its variants such as CatGAN \cite{springenberg2015unsupervised} is that we consider the diversity of instances in the minibatch sampling process. 
More specifically, for all instances in $X$, we partition them into a number of non-overlapping subspaces alike in certain features $\{X_{1}, \dots, X_{b}\}$ such that instances in the same subspace are more similar than those in different subspaces. 
Accordingly, labeled instances in $X^L$ and unlabeled instances in $X^U$ are partitioned into these $b$ subspaces, i.e., $X_i^L=X_i\cap X^L$ and $X_i^U=X_i\cap X^U$.

Let $\textbf{v} = (\textbf{v}_1, ..., \textbf{v}_b)$ be a vector corresponding to $b$ subspaces, where each $\textbf{v}_i=({v}^{1}_i,\dots, {v}^{n_i}_i)^T \in [0,1]^{n_i}$ and $n_i=|X^U_i|$. That is, each ${v}^{j}_i$ $(1\leq j\leq n_i)$ is associated with an instance in $X^U_i$. Then, a minibatch of $m$ instances is selected from $X^U$ according to the following objective function:
\begin{equation}
  \text{maximize} \qquad ||\textbf{v}||_{2,1} \quad \text{s.t.} \ \sum_{i,j} v^j_i= m
  \label{eq_div}
\end{equation}
where $||\textbf{v}||_{2,1}$ is a $l_{2,1}$-norm function defined as:
\begin{equation}
 ||\textbf{v}||_{2,1} = \sum^{b}_{i = 1} ||\textbf{v}_i||_{2} = {\sum^{b}_{i = 1}  \sqrt{\sum^{n_i} _{j=1} {v^j_i}^2}  }
\end{equation}
Here, $||\textbf{v}_i||_{2}$ is the $l_2$-norm of $\textbf{v}_i$. When ${v}^{j}_i=1$, the instance in $X^U_i$ corresponding to ${v}^{j}_i$ is selected into the minibatch; otherwise, that instance is not selected. 
When the value of the $l_{2,1}$-norm is small, instances are selected from a small number of subspaces in ${X}^U$ and the diversity of instances is low. Conversely, when maximizing the $l_{2,1}$-norm in Eq.~\ref{eq_div}, instances are selected from as many subspaces in $X^U$ as possible and the diversity of instances is high.

\medskip\noindent\textbf{Objective function of $G$.~} After a minibatch of unlabeled instances is selected from $X^U$ according to Eq.~\ref{eq_div}, the label generator $G$ generates a pseudo label $G(x_i)$ for each unlabeled instance $x_i$ in the minibatch.
Then, $(x_i,G(x_i))$ is sent to the discriminator $D$. After receiving the gradient from $D$, $G$ updates its parameters according to the following objective: 
\begin{equation}\begin{aligned}
  \mathcal{L}_G = &\mathop{min} \limits_{G} \quad  \mathbb{E}_{x \sim p(X^U_i)} 
  [\text{log}(1-D(x, G(x)))] 
\end{aligned}
\label{eq_G}
\end{equation}

\subsection{Discriminator}
\label{sec_d}
Unlike GAN, a discriminator $D$ in our \textsc{ErGAN} does not know about the real distribution $p(X,Y)$. Instead, $D$ has access only to a limited number of instances with real labels, i.e. $(X^L, Y)$. The goal of $D$ is to distinguish whether a labeled instance $(x,G(x))$ is from the real distribution $p(X,Y)$, i.e., given a pair $(x,G(X))$ as input, $D$ generates a scalar value in $[0,1]$ to indicate the probability that $G(x)$ is the same as the real label $y$ of $x$.  

\medskip\noindent\textbf{Propagation module.~} To achieve the above goal, as opposite to GAN and its variants in which the discriminator has the true distribution $p(X,Y)$, $D$ in \textsc{ErGAN} is designed to approximate the true joint distribution $p(X,Y)$ progressively through a propagation module. The general principle of propagation is that, the more confident the pseudo label $G(x)$ of an instance $x$ is the same as its real label $y$, the more likely such an instance is selected. Specifically, let $(X^{t}, G(X^{t}))$ denote all unlabeled instances with their pseudo labels at the $t$-th iteration of propagation. These instances are fed to $D$ to obtain their scores $D(X^{t}, G(X^{t}))$ that indicates the probabilities of their pseudo labels being the same as their real labels.
Based on the scores, a subset $\Delta X^t\subseteq X^t$ of instances is selected according to the following objective function:
\begin{equation}
\label{eq_prop}
\begin{multlined}
\argmax_{\Delta X^t \subseteq X^{t}} \sum_{x \in \Delta X^t} D(x, G(x)) \\ \text{   subject to  } |\Delta X^t| =\gamma
\end{multlined}
\end{equation}
where $\gamma$ is a hyper-parameter for the number of unlabeled instances being selected in the t-th iteration of propagation.

\begin{figure} 
		\centering
	\begin{center}
		\includegraphics[height = 0.24\textheight, width = 0.6\textwidth]{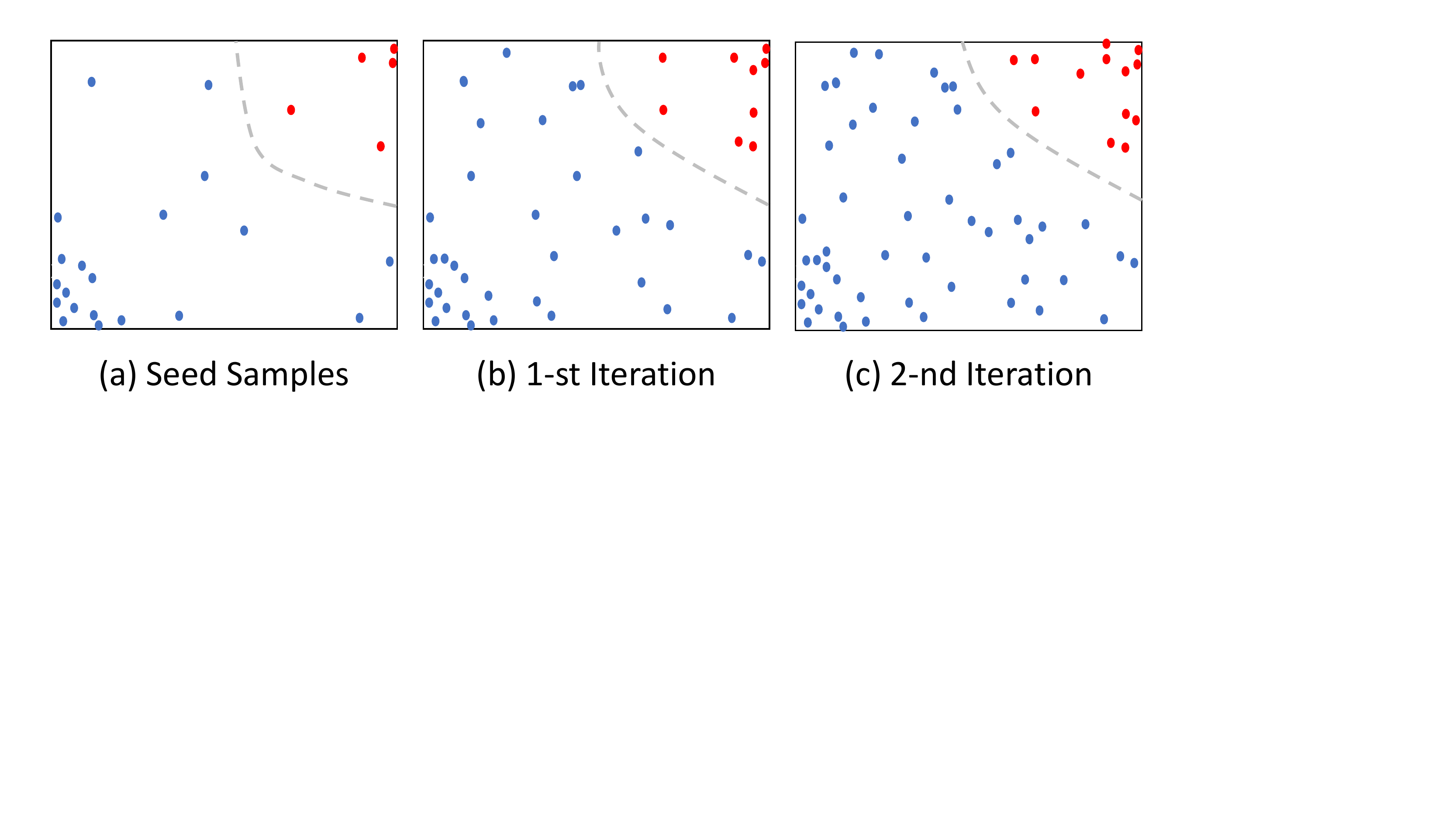}
		\vspace{-35mm}
	\end{center}
	\caption{\textbf{An illustration for propagation of \textsc{ErGAN}.} A boundary between two classes (red and blue) is learned through propagation.} \vspace{-2mm}
	\label{fig_samp}

\end{figure}

Then, this subset of instances with their high-quality pseudo labels $(\Delta X^t, \hat{Y})$ is propagated into the set of labeled instances $(X^*,Y)^{t}$ to train $D$, i.e.,
\begin{itemize}
    \item $(X^*,Y)^0=(X^L,Y)$
    \item $(X^*,Y)^{t}=(X^*,Y)^{t-1}\cup (\Delta X^t, \hat{Y})$
\end{itemize}
Hence, at the t-th iteration of propagation, $D$ has access to $(X^*,Y)^t$, which is a mixed set of labeled instances from $X^L$ (with real labels) and unlabeled instances from $X^U$ (with pseudo labels generated by $G$). The following holds:
\begin{equation}
(X^*,Y)^0\subseteq  (X^*,Y)^1\subseteq \dots \subseteq (X^*,Y)^t
\end{equation}
Figure~\ref{fig_samp} shows an example of the propagation in two iterations, where the grey dash line indicates a boundary between two classes (red and blue) and is learned through propagation.

\medskip\noindent\textbf{Objective function of $D$.~} The objective function of $D$ at the t-th iteration of propagation is defined as:
\begin{equation}
\begin{aligned}
  \mathcal{L}_D = &\mathop{max} \limits_{D} \quad  \mathbb{E}_{x \sim p(X^U_i)} \text{log}[(1-D(x, G(x)))] \\
  & + \lambda \mathbb{E}_{(x, y) \sim (X^*,Y)^t}\text{log}[D(x, y)] 
\end{aligned}
\label{eq_D}
\end{equation}
where $\lambda$ refers to a {weighted term}. 
In the following, we will explain how unlabeled instances with their pseudo labels, i.e., $(X^t, \hat{Y})$, is selected at the t-th iteration of propagation.

\subsection{Algorithm Description}

Our algorithm is described in Algorithm~\ref{Algo_ERGAN}. It involves two key processes: batch training (Lines 3-8) and label propagation (Lines 9-12). In the batch training process, a minibatch is randomly sampled from unlabeled instances in $X^U$ and $G$ generates pseudo labels for samples in this minibatch (Lines 4-5). Then another minibatch is randomly sampled from instances in $(X^*,Y)^t$ (Line 6). 
 After that, the discriminator $D$ is trained using these two minibatches w.r.t. Eq. \ref{eq_D}, while the label generator $G$ is trained w.r.t. Eq. \ref{eq_G}. The label propagation process only occurs after each batch training is finished. At this time, $G$ and $D$ reach (or closely approach) the equilibrium point and their parameters are fixed. In the label propagation process, $G$ first generates pseudo labels for all unlabeled instances in $X^t$, and then $D$ predicts scores for these instances. Based on these scores, a subset of instances $\Delta X^t\subseteq X^t$ with high-quality pseudo labels is propagated into the set of labeled instances $(X^*, Y)^t$ for $D$ in the t-th iteration.


\medskip
\noindent\textbf{How to choose $b$, $t$ and $n$?} In our algorithm, the number of subspaces $b$ is decided based on the attributes in each dataset. Suppose that a dataset has four attributes, we first obtain the median value for each attribute, and then partition instances into $4^2=16$ subspaces according to whether attribute values of each instance are above or below the median values of these four attributes \cite{shao2019learning}. 
Furthermore, $n$ is a hyper-parameter referring to the number of iterations for converging $G$ and $D$, and $t$ is decided by the total number $X^U$ of unlabeled instances and the number $\gamma$ of instances being propagated in each iteration, i.e. $t = \lceil \frac{|X^U|}{\gamma} \rceil$.


\begin{algorithm}
	\KwIn{ $b$ subspaces in $X$; ${X}^U$; $(X^L, Y)$;\\
	}
	\KwOut{
		 $(X^*, Y)^t$ where $X^*=X$
	}	
    Initialize $t = 0$; $(X^*,Y)^0=(X^L,Y)$; $X^0 = X^1=X^U$\\
    \While {${X}^t \neq \emptyset$}{
    \For (\tcp*[f]{Batch training}){n \emph{iterations}}{ 
    Sample a minibatch $\{x_1, ..., x_m\}$ from $X^{U}$ w.r.t. Eq.~\ref{eq_div}  \\
    Generate pseudo labels $\{(x_1, G(x_1)),..., (x_m, G(x_m))\}$\\
     Sample a minibatch $\{(x^L_1, y_1), ..., (x^L_m, y_m)\}$ from $(X^*,Y)^t$\\   
    Update the parameters of $D$ w.r.t. Eq.~\ref{eq_D} \\ %
    Update the parameters of $G$ w.r.t. Eq.~\ref{eq_G} \\ 
    }
    t=t+1 \tcp*[f]{Label propagation}\\ 
    Generate pseudo labels for $X^t$ \\
    Select $\Delta X^t\subseteq X^t$ for propagation w.r.t. Eq.~\ref{eq_prop}\\ 
   $(X^*,Y)^{t}=(X^*,Y)^{t-1}\cup (\Delta X^t, \hat{Y})$;  ${X}^{t+1} = {X}^t -\Delta X^t$    }
	\caption{Minibatch stochastic gradient descent and label propagation of \textsc{ErGAN}}
	\label{Algo_ERGAN}	

\end{algorithm}

\section{Theoretical Analysis}
We prove several nice theoretical properties of \textsc{ErGAN}.

\smallskip
\noindent\textbf{Diversity.~} We first show that partitioning $X$ into subspaces for diversity does not affect the learning goals of $G$ and $D$.

\begin{lemma}
\label{lem_subg}
  If $p_g(Y|X_i^U)$ by $G$ approximates $p(Y|X_i^U)$ for each feature subspace $X_i^U$, then $p_g(Y|X^U)\approx p(Y|X^U)$. 
\end{lemma}
\begin{proof} $X^U$ is partitioned into non-overlapping subspaces $\{X^U_i\}_{i=1}^b$. Thus, $p(X^U)$ is composed of $b$ distributions $p(X^U_i) (1\leq i\leq b)$. Further, Eq.~\ref{eq_div} prefers instances from different subspaces, but within each subspace, instances are selected randomly. 
\end{proof}

Since unlabeled and labeled instances are sampled from the corresponding subspaces, we have the following lemma for $D$.
\begin{lemma}
\label{lem_subd}
  If $p_d(X_i^U, Y)$ by $D$ approximates $p(X_i^U,Y)$ for each feature subspace $X_i^U$, then $p_d(X^U,Y)\approx p(X^U,Y)$. 
\end{lemma}
By $|X^L| <\!< |X^U|$, we have $p(X^U)\approx p(X)$. Accordingly, $p_g(Y|X^U)$ $\approx p_g(Y|X)$, $p_d(X^U,Y)\approx p_d(X,Y)$, and $p(X^U,Y)\approx p(X,Y)$. 

\medskip\smallskip
\noindent\textbf{Global optimality.~} 
Here we use $p_d(X^*, Y)^t$ to refer to the distribution learned by $D$ at the t-th iteration of propagation. We have the following lemma.
\begin{lemma}
  For the fixed label generator $G$, the optima $D$ in the t-th iteration of propagation is:
  \begin{equation}
  D^*_G(x,y) = \frac{p_d(X^*, Y)^t}{p_d(X^*, Y)^t + p_g(X, Y)^t}
  \end{equation}
  \label{lem_D}
\end{lemma}
\begin{proof} At the t-th iteration of propagation, by Eq.~\ref{eq_D}, the training objective of $D$ is to maximize:
\begin{equation*}
\begin{aligned}
    & \mathbb{E}_{x \sim p(X^U_i)} \text{log}[(1-D(x, G(x)))] +  \\
    &\hspace{1.5cm} \lambda \mathbb{E}_{(x, y) \sim p_d(X^*, Y)^t}\text{log}[D(x, y)] \\
\end{aligned}
\label{eq_o1}
\end{equation*}
Based on the "change of variable" technique \cite{gardiner1985handbook}, we have $p_g(s) = p_{X^U_i}(G^{-1}(y))\frac{dG^{-1}(y)}{d(s)}$, where $s=(x,y)$ denotes the vector concatenated by $x$ and $y$ at the t-th iteration of propagation. The first part of the above formula equals to:
\begin{equation*}
\begin{aligned}
     & \int_{s} p_{X^U_i} (G^{-1}(y)) \text{log} (1-D(s)) dG^{-1}(y)\\
    = & \int_{s} p_g(s) \text{log}[(1-D(s))]d(s) \\
\end{aligned}
\end{equation*}

Hence, we have an objective function that is the same as GAN \cite{goodfellow2014generative}. This lemma is proven.
\end{proof}
As a result, we have the following lemma.
\begin{lemma}\label{lem_G}
  The global minimum of the training criterion of $G$ under the optimal $D$ at the t-th iteration of propagation is achieved when $p_d(X^*,Y)^t = p_g(X, Y)^t$.
\end{lemma}

\smallskip
\noindent\textbf{Equilibrium.~} 
By Lemma \ref{lem_G}, we have $p_g(X,Y)^t=p_d(X^*,Y)^t$. Thus, the equilibrium of the minimax game in \textsc{ErGAN} is achieved at each iteration of propagation. When the labels are propagated sufficiently, the real data distribution is well simulated. The following lemma states that $D$ can approximate the real distribution $p(X,Y)$ when the number of propagation iterations is large. 
\begin{lemma}
    $p_d(X^*,Y)^t$ approaches $p(X,Y)$ when $t$ increases.
\end{lemma}
Statistically, according to the \emph{Central Limit Theorem}, the larger sample size the more likely an estimated distribution is close to the real distribution. This lemma corresponds to the central property of self-training based semi-supervised learning approaches \cite{grandvalet2005semi,triguero2015self}.

\medskip
\noindent\textbf{Mode collapse.~} The mode collapse problem occurs in GAN where the generator collapses to generate limited and nonsensical images. However, our approach \textsc{ErGAN} does not suffer from this problem for two reasons:
(1) \textsc{ErGAN} takes unlabeled instances as input for both $G$ and $D$. This allows $D$ to distinguish $p_d(X^*,Y)^t$ from $p_g(X,Y)^t$, and $D$ can prevent $G$ to generate the same kind of pseudo labels for unlabeled instances, i.e., mode collapse.
(2) In GAN, mode collapse only occurs when the discriminator is well trained but the generator is not optimal. In \textsc{ErGAN},  $(X^*,Y)^t$ only has limited instances with labels when $t$ is small. Thus, $D$ can hardly be optimal in early iterations of propagation. For later iterations, since $G$ is trained iteratively based on its previous parameters, the mode collapse issue can also be avoided.

\section{Experimental Setup}

We evaluate \textsc{ErGAN} to answer the following questions: 
\begin{itemize}
    \item[\textbf{Q1.}] 
    How does \textsc{ErGAN} perform in comparison with the state-of-the-art unsupervised, semi-supervised and fully supervised methods? 

    \item[\textbf{Q2.}] How do the design choices such as the diversity module, the propagation module, and GAN architecture affect performance of \textsc{ErGAN}?  
    
        \item[\textbf{Q3.}] To what degree  \textsc{ErGAN} can work for classifying instances when the label cost is extremely limited (i.e. only a small number of instances with real labels)? 
\end{itemize}

\smallskip
\noindent\textbf{Datasets.~} We use four widely used benchmark datasets of ER tasks: (1) \emph{Cora}\footnote[1]{Available from: \emph{http://secondstring.sourceforge.net}} dataset contains bibliographic records of machine learning publications. (2) \emph{DBLP-Scholar}\footnotemark[1] dataset contains bibliographic records from the DBLP and Google Scholar websites. (3) \emph{DBLP-ACM} \cite{kopcke2010evaluation} dataset contains bibliographic records from the DBLP and ACM websites.
(4) \emph{North Carolina Voter Registration (NCVoter)}\footnote[2]{Available from: \emph{http://alt.ncsbe.gov/data/}} dataset contains real-world voter registration information of people from North Carolina in the USA. Table~\ref{tab_db} summarizes the characteristics of these four datasets. We can see that the datasets are highly imbalanced, i.e., the number of instances from the majority class (non-match) is much more than the number of those from the minority class (match).

\begin{table}[t!]
	\caption{\textbf{Characteristics of datasets.} The instances of these datasets are generated from their record pairs.}\vspace{-0.1cm}\label{tab_db}
	\centering	
	\resizebox{0.48\textwidth}{!}{\begin{tabular}{|l|cccc|}
		\midrule
		\multirow{2}{*}{\textbf{Dataset}} & \textbf{$\#$Attributes} & {\textbf{$\#$Instances} } & \textbf{Imbalance} & \textbf{\#Subspaces}\\ &($|A|$)&($|X|$)& \textbf{Rate} & ($b$)\\
        \midrule
		Cora & 4 & 837,865 &1:49 &16\\ \cline{1-1}
		DBLP- & \multirow{2}{*}{4/4} &  \multirow{2}{*}{6,001,104} & \multirow{2}{*}{1:2,698} &\multirow{2}{*}{16}\\	
		ACM&&&&\\\cline{1-1}
		DBLP- & \multirow{2}{*}{4/4} &  \multirow{2}{*}{168,112,008} & \multirow{2}{*}{1:71,233} &\multirow{2}{*}{16}\\
		Scholar&&&&\\\cline{1-1}
		NCVoter & 18/18 &  1,000,000 & 1:4,202 &64\\
		\midrule
	\end{tabular}}
		\vspace{-3mm}
\end{table}


\medskip
\noindent\textbf{Baselines.~} We compare \textsc{ErGAN} with the following baselines: 
(1) \emph{Unsupervised methods:} \textbf{Two-Steps} (\textbf{2S}) is a widely-used unsupervised learning method proposed by Christen \cite{christen2008automatic}.  \textbf{Iterative Term-Entity Ranking and CliqueRank} (\textbf{ITER-CR}) is the state-of-the-art graph based unsupervised method for ER \cite{zhang2018graph}. 
(2) \emph{Semi-supervised methods:} \textbf{Semi-supervised Boosted Classifier} (\textbf{SBC}) is the state-of-the-art semi-supervised learning method with label propagation based on Adaboost classifier \cite{ratsch2001soft} proposed by Kejriwal and Miranker \cite{kejriwal2015semi}. 
(3) \emph{Fully supervised methods:} \textbf{Magellan} is a state-of-the-art open-source fully supervised learning-based ER solution designed by Konda et. al. \cite{konda2016magellan}. We consider two supervised classifiers provided in {Magellan}: \textbf{Logistic Regression (LR)} and \textbf{Support Vector Machine (SVM)}.
\textbf{eXtreme Gradient boosting} (\textbf{XGboost}) is a state-of-the-art fully supervised ensemble learning based method proposed by Chen and Guestrin \cite{chen2016xgboost}. \textbf{DeepMatcher} (\textbf{DM}) is a state-of-the-art deep learning based entity matching method for ER \cite{mudgal2018deep}. \textbf{Deep Transfer active learning} (\textbf{DTAL}) is the state-of-the-art active learning method which combines both transfer learning and active learning for handling ER tasks \cite{kasai2019low}.

To make a fair comparison, we follow the default parameters suggested in the original papers of the baselines. Note that DM uses the imbalance rate as a hyper-parameter which is normally unknown in real-life applications. Both DTAL and DM are deep learning based methods in which FastText \cite{bojanowski2017enriching} is used for learning word embeddings. For other baselines, we use 2-gram Jaccard similarity for textual comparison.


To compare with the baselines that use word embeddings, we use \textbf{\textsc{ErGAN+WE}} to refer to the model of \textsc{ErGAN} augmented with word embeddings for attribute values. In our ablation study, we use \textbf{\textsc{ErGAN-D}} and \textbf{\textsc{ErGAN-P}} to refer to a model being obtained by removing the diversity and propagation modules from \textsc{ErGAN}, respectively, and \textbf{\textsc{ErNN}} a model in which the GAN architecture (i.e. $G$ and $D$ are trained alternatively) is replaced by a single multi-layer perceptron for semi-supervised learning with the diversity module. We set $\lambda=1$, $m\leq 100$, and $\gamma = |X^*|$ at each iteration of propagation. Our models use the same word embedding and similarity comparison techniques as the baselines.   


\medskip
\noindent\textbf{Measures.~} We use the following widely used measures in ER tasks for performance evaluation \cite{christen2012data}: (1) \emph{Recall} (R) is the fraction of true matches being predicted by the classifier among the total number of true matches; (2) \emph{Precision} (P) is the fraction of true matches over all matches being predicted by the classifier; (3) \emph{F-measure} (FM) is the harmonic mean of recall and precision, i.e. $\emph{FM} = \frac{2*R*P}{R + P}$. Due to the existence of highly imbalanced classes in ER datasets, F-measure is preferred rather than accuracy in our experiments.

\section{Experimental Results}

In this section, we discuss the results of our experiments to answer the aforementioned questions. 

\begin{table}[t!]
\caption{\textbf{Experimental results of f-measure with 60\% training.} The results marked by $*$ are taken from the original papers and the others are obtained by running the code provided by the authors.}\vspace{-0.1cm}
	\centering
	\label{tab_fmful}
	\begin{tabular}{|l|cccc|}\midrule
        \multirow{3}{*}{\hspace{0.3cm}\textbf{Method}} &\multicolumn{4}{c|}{\textbf{Datasets}}\\\cline{2-5} &\multirow{2}{*}{Cora} & DBLP- & DBLP- & \multirow{2}{*}{NCVoter}\\
        & & ACM&Scholar &\\
        \midrule
        2S \cite{christen2008automatic} & 62.69 & 91.43 & 68.78 & 98.96\\
        ITER-CR* \cite{zhang2018graph} & 89.00 & -- & -- & --\\
        SBC \cite{kejriwal2015semi} & 85.71 & 97.09 & 85.47 & 99.78\\
        SVM \cite{konda2016magellan}& 88.95 & 97.19 & 85.71 & 98.48 \\
        LR \cite{konda2016magellan} & 80.25 & 95.56 & 83.84 & 99.37 \\
        XGBoost \cite{chen2016xgboost}& 91.34 & 97.20 & 86.63 & \textbf{100} \\
        \textsc{ErGAN} & \textbf{93.03} & \textbf{98.23} & \textbf{88.32} & \textbf{100} \\
        \midrule
        DM \cite{mudgal2018deep} & 98.58 & 98.29 & 94.68 & \textbf{100}\\
        DTAL* \cite{kasai2019low} & $98.68_{\pm0.26}$ & $98.45_{\pm0.22}$ & $92.94_{\pm0.47}$ & --\\
        \textsc{ErGAN+WE} & \textbf{98.72}$_{\pm0.15}$ & \textbf{98.51}$_{\pm0.23}$ & \textbf{94.73}$_{\pm0.35}$ & \textbf{100} \\
        \bottomrule
        
    \end{tabular}
\end{table}

\begin{figure*} 
	\begin{center}
		\includegraphics[height = 0.18\textheight, width = 0.98\textwidth] {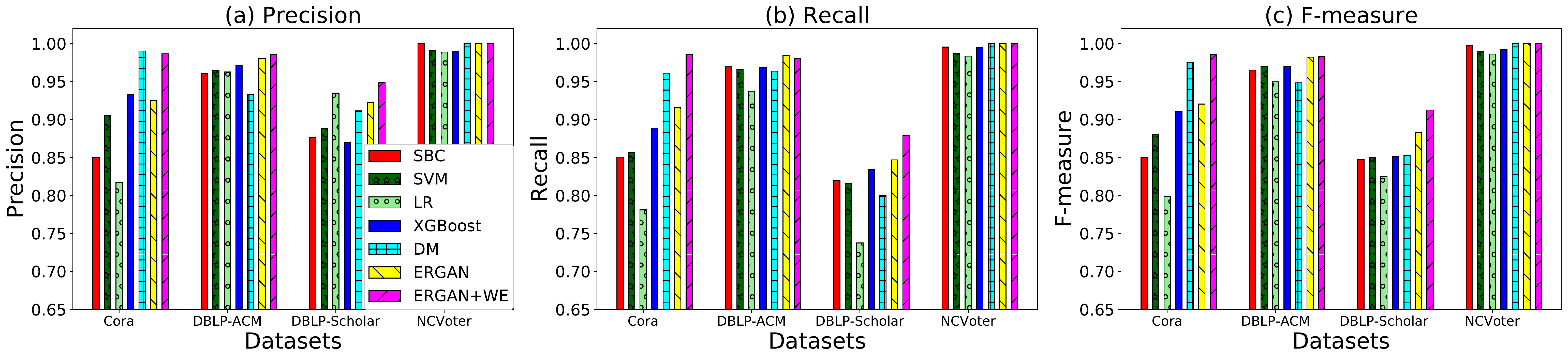}
\vspace{-0cm}
	\caption{\textbf{Experimental results of precision, recall and f-measure with 20\% training.}} 
	\label{fig_bar}\vspace{0.2cm}
		\includegraphics[height = 0.2\textheight, width = \textwidth]{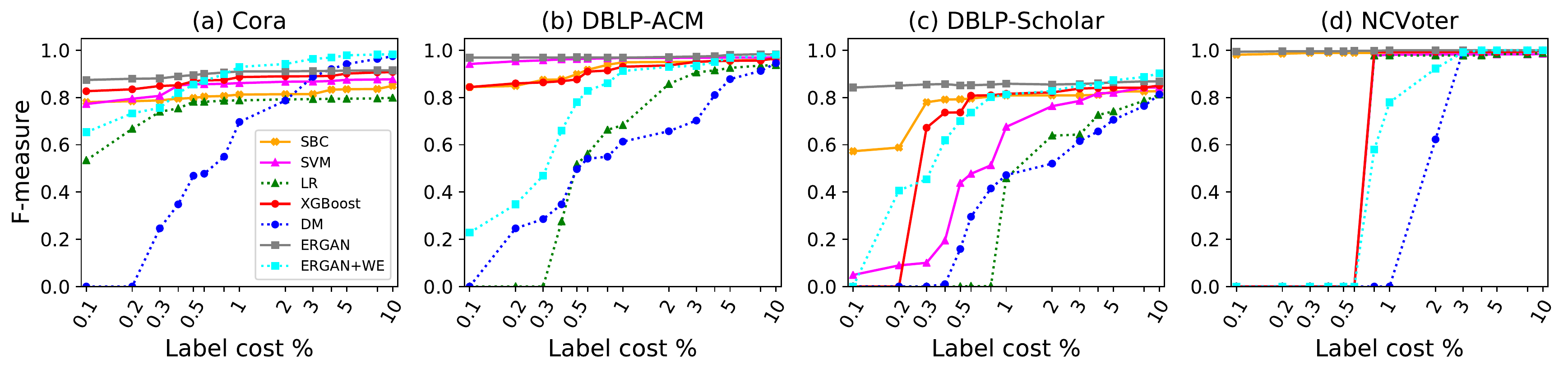}
	\end{center}\vspace{-0.3cm}
	\caption{\textbf{Experimental results of f-measure with 0.1\% -- 10\% training.}} 
	\label{fig_fm}
\end{figure*}

\subsection{Performance Comparison} 


\smallskip
\noindent\textbf{Task 1.} We conduct an experiment to evaluate how our methods perform against the baselines. Following the previous work for the supervised methods DM \cite{mudgal2018deep} and DTAL \cite{kasai2019low}, we split the datasets with 60\% for training and the rest for testing.  

Table~\ref{tab_fmful} shows the results of the experiment, where the last three methods DM, DTAL and \textsc{ErGAN+WE} are deep-learning methods which use word embeddings for attribute values and the other methods use Jaccard similarity for comparing attribute values (without using word embeddings). 
We can see that, the unsupervised method 2S performs the worst among all the methods. However, the other unsupervised method ITER-CR performs better than SBC, SVM and LR due to its ability to leverage graph based structure. Compared with the fully supervised methods, the semi-supervised method SBC performs better than LR, comparably with SVM, but worse than XGBoost. Our method \textsc{ErGAN} performs better than any non-deep-learning method, but worse than the deep-learning methods with word embedding, i.e., DM, DTAL and \textsc{ErGAN+WE}. Nonetheless, our method \textsc{ErGAN+WE} outperforms all the baseline, including two deep-learning methods DM and DTAL, over all databases they have the results.

\medskip
\noindent\textbf{Observation 1.} \emph{With sufficient training data, \textsc{ErGAN+WE} performs better than \textsc{ErGAN} due to the power of word embedding for records. \textsc{ErGAN+WE} has superior performance against all the baselines consistently.}

\medskip
\noindent\textbf{Task 2.} To understand how performance may change when the amount of training data is reduced, we conduct an experiment on the supervised methods by splitting the datasets into 20\% for training and 80\% for testing. DTAL is excluded in this experiment because the original paper does not have results in this setting and also no code is available. 

Figure~\ref{fig_bar} shows the results of precision, recall and f-measure over four datasets. 
LR performs the worst except for precision on DBLP-ACM. The performance of the semi-supervised method SBC is better than LR and SVM, and comparable with XGBoost w.r.t. f-measure. This is because SBC takes the advantage of label propagation but its classifier Adaboost is not as powerful as XGBoost. 
For the deep-learning methods, the performance of DM drops significantly on DBLP-ACM and DBLP-Scholar and even worse than \textsc{ErGAN} which is different from the case of 60\% training in Table \ref{tab_fmful}. \textsc{ErGAN+WE} still performs the best on all datasets w.r.t. all the three measures. Overall, compared with Table~\ref{tab_fmful}, the performance of SVM, LR and DM is affected considerably on one or more datasets, whereas SBC, XGBoost,  \textsc{ErGAN} and \textsc{ErGAN+WE} remain
comparable performance.





\medskip
\noindent\textbf{Observation 2.}
\emph{With reduced but still considerable training data, the performance of \textsc{ErGAN} and \textsc{ErGAN+WE} remains strong and consistent. This is because \textsc{ErGAN} and \textsc{ErGAN+WE} benefit from its adversarial learning architecture and diversity and propagation modules.}


\medskip
\noindent\textbf{Task 3.}  To study performance under a limited number of instances with real labels, we further conduct an experiment using only a small percentage for training, ranging from 0.1\% to 10\% of the datasets, and the rest for testing. 

Figure~\ref{fig_fm} shows the experimental results. \textsc{ErGAN} performs best among all the methods over all the datasets when training data is below 1\%. \textsc{ErGAN+WE} performs poorly in this range. However, the performance of \textsc{ErGAN+WE} increases rapidly with increasing training data and exceeds all the other methods on all the datasets when training data reaches 10\%. LR and DM have a similar trend as \textsc{ErGAN+WE}, but perform significantly worse. The semi-supervised method SBC performs better than \textsc{ErGAN+WE} only when training data is small, i.e. below 0.2\% for Cora and below 0.9\% for DBLP-ACM, DBLP-Scholar and NCVoter. The performance of SVM and XGBoost varies in datasets, i.e., perform well on Cora and DBLP-ACM, but badly on DBLP-Schloar and NCVoter. This demonstrates that the performance of SVM and XGBoost is sensitive to the imbalance rate of a dataset, and they fail to handle imbalanced data when no sufficient training data is available.
For NCVoter, due to a clear boundary existing between matches and non-matches in the underlying distribution, the performance of all the methods that perform poorly for small training data can be dramatically improved after using 0.6\% or more training data. In general, we may conclude that, compared with the cases of 60\% and 20\% training in Table~\ref{tab_fmful} and Figure~\ref{fig_bar}, the performance gain of the methods with word embedding against the methods without word embedding does not exist anymore. Instead, the methods with word embedding performs worse than most of the methods without word embedding when training data is small, i.e., below 1\%.

\medskip
\noindent\textbf{Observation 3.} \emph{When decreasing training data, \textsc{ErGAN+WE} gradually performs worse than \textsc{ErGAN+WE}. This is because, \textsc{ErGAN+WE} transforms instances into a high dimensional space through word embedding and thus requires much more labels in training than \textsc{ErGAN}.}

\begin{table*}[ht!]
\caption{\textbf{Experimental results of f-measure with 0.1\%, 1\%, 20\% and 60\% training for ablation analysis.}}\vspace{-0cm}
	\centering
	\label{tab_abl}
	\scalebox{0.98}{
	\begin{tabular}{|l||cccc|cccc|cccc|cccc|}
        \hline
        \multicolumn{1}{|c||}{\multirow{2}{*}{\textbf{Datasets}}} & \multicolumn{4}{|c|}{\textbf{Cora}} & \multicolumn{4}{|c|}{\textbf{DBLP-ACM}} & \multicolumn{4}{|c|}{\textbf{DBLP-Scholar}} & \multicolumn{4}{|c|}{\textbf{NCVoter}}\\  
        \cline{2-17}
        & 0.1\% & 1\% & 20\% & 60\% & 0.1\% & 1\% & 20\% & 60\% & 0.1\% & 1\% & 20\% & 60\% & 0.1\% & 1\% & 20\% & 60\% \\        \hline
        
        \textsc{ErNN}    & 84.46 & 90.67 & 91.43 & 92.78 & 88.05 & 95.68 & 98.20 & 98.22 & 82.76 & 83.17 & 86.71 & 87.73 & 99.39   & 100   & 100   & 100\\ 
        
        \textsc{ErGAN-D} & 79.87 & 85.14 & 91.27 & 92.97 & 0     & 93.30 & 97.16 & 98.21 & 0     & 78.85 & 83.43 & 88.29 & 0       & 99.58 & 100   & 100\\ 
        
        \textsc{ErGAN-P} & 85.18 & 90.76 & 91.42 & 93.03 & 92.67 & 95.96 & 98.21 & 98.23 & 83.43 & 85.34 & 86.55 & 88.32 & 99.39   & 99.79 & 100   & 100\\ 
        
        \textsc{ErGAN}   & 87.45 & 91.07 & 91.54 & 93.03 & 96.89 & 96.93 & 98.22 & 98.23 & 84.23 & 85.85 & 86.86 & 88.32 & 99.45   & 100   & 100   & 100 \\

		\hline

	\end{tabular}
    }
\end{table*}

\begin{table}[t]
\caption{\textbf{Experimental results of f-measure under extremely limited real-labeled instances.} The methods SVM, LR, XGBoost, DM, \textsc{ErGAN-D}, and \textsc{ErGAN+WE} have the f-measure value 0 in all these settings and are thus excluded from the table.}\vspace{-0cm}
	\centering
	\label{tab_rob}
	\scalebox{1.0}{
	\begin{tabular}{c|c|cccc}
        \midrule
        \multicolumn{1}{c|}{\multirow{2}{*}{\textbf{Dataset}}} & \textbf{Label Cost} &\multicolumn{4}{|c}{\textbf{Methods}}\\\cline{3-6}
         & (\textbf{\#Instances})  & {SBC} & {\textsc{ErNN}} & {\textsc{ErGAN-P}} & {\textsc{ErGAN}}  \\  
        \midrule
        
        \multirow{4}{*}{Cora}&50  & 0 & 0.6648 & 0.7358 & \textbf{0.7735} \\ 
        
        &100&  0 & 0.7684 & 0.7960 &\textbf{0.8083}\\ 
        
        &200 & 0.303 & 0.7742 & 0.8156&\textbf{0.8314} \\ 
        
        &500 & 0.7629 & 0.8234 & 0.8493 &\textbf{0.8691} \\  
		\midrule
     \multirow{4}{*}{}&50  & 0 & 0.6694 & 0.8492 &\textbf{0.8869} \\ 

        DBLP-&100&  0 & 0.7261 & 0.9143 &\textbf{0.9673} \\ 
        
        ACM&200&  0 & 0.7463 & 0.9151 &\textbf{0.9656} \\ 
        
        &500& 0 &  0.8013 & 0.9174 & \textbf{0.9681} \\
		\midrule

        \multirow{4}{*}{}&50 & 0 & 0.0043 & 0.6777 &\textbf{0.7760} \\ 

        DBLP-&100 & 0 & 0.0536 & 0.7335 &\textbf{0.8045} \\ 
        
        Scholar&200 & 0 & 0.6869 & 0.7869 &\textbf{0.8124} \\ 
        
        &500 & 0 & 0.7903 & 0.8256 &\textbf{0.8372} \\ 
		\midrule
        \multirow{4}{*}{NCVoter} &
        50  & 0 & 0.3603 & 0.6389 &\textbf{0.7192} \\ 
        &100  & 0 & 0.8202 & 0.9091&\textbf{0.9532} \\ 
        
        &200 &  0 & 0.9289 & 0.9431&\textbf{0.9583}\\

        &500&  0 & 0.9724 & \textbf{0.9740} &\textbf{0.9740} \\ 
        
        
		\midrule
	\end{tabular}
    }\vspace{-2mm}
\end{table}

\subsection{Ablation Analysis}

We conduct an ablation study to evaluate the effects of the key components of \textsc{ErGAN}, including the adversarial learning architecture, the diversity module and the propagation module, under different label costs, ranging from 0.1\% to 60\% for training. The results are presented in Table~\ref{tab_abl}.
We observe that the performance of all the methods \textsc{ErNN}, \textsc{ErGAN-D}, \textsc{ErGAN-P} and \textsc{ErGAN} become stable and gradually converge when the label cost increases, e.g. in the case of 60\% training. Nonetheless, \textsc{ErGAN} performs the best among all the methods, and the performance of the other methods varies in different datasets. 
In the following, we will discuss how each key component of \textsc{ErGAN} may affect the performance. 
 
\medskip
\noindent\emph{Adversarial learning architecture.} The performance of \textsc{ErNN} generally lies in between \textsc{ErGAN-D} and \textsc{ErGAN-P}, and significantly worse than \textsc{ErGAN}. This indicates that the use of adversarial learning architecture by \textsc{ErGAN} helps to improve the performance, particularly when training data is limited, e.g., for 0.1\% training, \textsc{ErGAN} improves around 3\% on Cora and more than 8\% on DBLP-ACM upon \textsc{ErNN}.

\medskip
\noindent\emph{Diversity module.} In Table~\ref{tab_abl}, the results of \textsc{ErGAN-D} are the worst among all the methods over all the datasets. This indicates that diverse instances are more informative for model training, which can improve the label efficiency. Specifically, with 0.1\% training, \textsc{ErGAN-D} fails to work (i.e., f-measure value is 0) on three datasets except for Cora. This is because \textsc{ErGAN-D} lacks the diversity module and can only randomly select instances for training. As a result, all training instances are selected from the majority class (non-matches), and accordingly no matched instance can be classified correctly by \textsc{ErGAN-D}, i.e. all the instances are classified as non-matches. Since datasets in ER applications are usually highly imbalanced, training data without diversity may hardly contain instances from the minority class (matches) when labels are limited, thus leading to poor performance.

\medskip
\noindent\emph{Propagation module.} Table~\ref{tab_abl} shows that \textsc{ErGAN-P} generally has better performance than \textsc{ErNN} and \textsc{ErGAN-D}, and thus it may affect the performance of \textsc{ErGAN} least compared with the other two key components: the adversarial learning architecture and the diversity module, especially when the label cost is small, e.g. 0.1\% and 1\% training. Additionally, when the label cost is 60\%, the performance of \textsc{ErGAN-P} and \textsc{ErGAN} is the same. This is because instances with real labels in 60\% training data can provide sufficient information for learning, and the propagation of instances with pseudo labels becomes unnecessary.





\medskip
\noindent\textbf{Observation 4.} \emph{In \textsc{ErGAN}, all the three key components, i.e., the adversarial learning architecture, the diversity module and the propagation module, are necessary, each serving as an integral part of the entire framework.}

\subsection{Extremeness Test}
In the previous experiments, \textsc{ErGAN} has demonstrated strong and consistent performance when training data is reduced from 60\% to 0.1\%. However, a question left is: what are the minimum label costs required by \textsc{ErGAN} to achieve reasonably performance? To answer this, we conduct an experiment under extremely limited label cost, ranging from 50 to 500 instances with real labels. 
Table~\ref{tab_rob} shows the results of our experiment. Note that we exclude the results of the methods SVM, LR, XGBoost, DM, \textsc{ErNN}, \textsc{ErGAN-D} and \textsc{ErGAN+WE} from this table because they fail to work when the label costs are below 500, i.e., f-measure value is 0 in all the settings in Table~\ref{tab_rob}. 

In Table~\ref{tab_rob}, the results of SBC are almost 0 except for the cases when the label cost is 200 and 500 on Cora. This shows that SBC as a semi-supervised method can perform better than other fully supervised methods SVM, LR, XGBoost and DM under extremely limited label cost. Moreover, the performance of SBC is affected by the imbalance rate, and SBC fails to perform when the imbalance rate of a dataset is high.

 We also notice that \textsc{ErGAN} can perform reasonably well on all the datasets even with 50 labels, and achieve good performance using only 500 labels. Moreover, \textsc{ErNN}, \textsc{ErGAN-P} and \textsc{ErGAN} have results in all the setting, but \textsc{ErGAN-D} does not have results in any of these settings. This is due to the diversity module, which can effectively select balanced training data to maximize the use of labels under extremely limited label cost. \textsc{ErGAN-P} performs better than \textsc{ErNN} on all the datasets. It indicates that performance may be harmed rather than helped by propagation if the quality of pseudo labels being propagated is not guaranteed. It is worthy to note that, for the cases of 50 and 100 labels on DBLP-Scholar dataset, we find that the reason why \textsc{ErNN} has very low feature values is because of high recall values (i.e., 0.749 for 50 labels and 0.777 for 100 labels) but low precision values (i.e., 0.002 for 50 labels and 0.028 for 100 labels). It means \textsc{ErNN} incorrectly classifies many non-matches as matches (i.e., false positives) and then propagates them into training data, leading to poor performance. With the increase in the label cost, this phenomenon is alleviated. 



\medskip
\noindent\textbf{Observation 5.} \emph{\textsc{ErGAN} can achieve good performance even with extremely limited labels. This is because the label generator $G$ and the discriminator $D$ are trained adversarially in \textsc{ErGAN} such that $G$ uses the diversity module to balance the selection of instances from different classes and $D$ propagates instances with high-quality psuedo labels into training.}

\section{Related Work}

\subsection{Entity Resolution}
{Entity Resolution} (ER) has been extensively studied for decades since it was first reported in 1946 \cite{dunn1946record,christen2012data,christophides2019end}. Traditionally, an ER task is performed through two stages (1) \emph{blocking}, and (2) \emph{matching}. Blocking aims to reduce the search space for record pair comparison which measures similarity of record pairs. Matching aims to determine whether record pairs refer the same real-world entity, and classification is the core problem in this stage. In this work, we focus on the classification problem of ER tasks. 

Generally, two kinds of classification approaches are widely used in ER: rule-based and learning-based. Rule-based classification approaches \cite{whang2010entity, fan2009reasoning}, often involve hand-crafted rules that associate with certain thresholds for defining similarity or dissimilarity of records \cite{christen2009development, singh2017synthesizing}. Learning-based approaches usually adopt a learning model to classify whether two records refer to the same entity. In the literature, there are three main categories: (1) Supervised learning approaches, which train a model to fit labeled instances so that it can predict unlabeled instances. The most recent work is \emph{Magellan} \cite{konda2016magellan}, which considered learning models including \emph{Decision Tree}, \emph{Random Forest} and \emph{Support Vector Machine} (SVM). Some work also studied ensemble learning approaches for ER to build a strong learned based on a set of weak learners \cite{freund1995boosting}. A widely used ensemble classifier is \emph{extreme gradient boosting} (XGBoost) \cite{chen2016xgboost}, which uses the sparsity-aware algorithm and the weighted quantile sketch for approximate learning. (2) Unsupervised learning approaches, which take no real labeled instances, and assign labels to instances based on prior knowledge \cite{bhattacharya2006latent, jurek2017novel}. A standard approach for ER is called \emph{two-steps} (2S) \cite{christen2008automatic}. It first labels a number (e.g. 10 percents of a dataset) of the most similar and dissimilar record pairs, respectively, and then trains a SVM in the second step. A most recent work in this line was proposed by Jurek et. al. \cite{jurek2017novel}, which considered both ensemble learning and automatic self-learning for classification based on training labels which are automatically generated based on different similarity measure schemes. 
A recent unsupervised approach for ER is proposed by Zhang et al. \cite{zhang2018graph} which has two components: Iterative Term-Entity Ranking (ITER) and CliqueRank for record graph construction. 
(3) Semi-supervised learning approaches, which sit between supervised and unsupervised learning in that they take both limited real labeled and unlabeled instances for training. The state-of-the-art semi-supervised learning approach is an ensemble learning based approach using ensemble classifier Adaboost \cite{ratsch2001soft} for label prediction based on seed instances that have real labels. 

\subsection{Generative Adversarial Network}
Generative adversarial network (GAN) was proposed by Goodfellow et. al. \cite{goodfellow2014generative}. The key idea of GAN is that  two networks, a \emph{generator} and a \emph{discriminator}, play a minimax game so that they converge gradually to an optimal solution. The generator aims to generate fake instances to "fool" the discriminator by simulating the distribution of real instances, while the discriminator targets to distinguish fake instances (generated by the generator) from real instances. Due to the success of GAN in generating realistic images, a large number of studies have extended GAN to dealing with various tasks such as instance classification with semi-supervised learning \cite{springenberg2015unsupervised, salimans2016improved}, labeled instance generation \cite{mirza2014conditional} and  label generation \cite{deng2018adversarial}. Various techniques have also been proposed to improve GAN's performance by alleviating the mode collapse and convergence problems \cite{salimans2016improved, arjovsky2017wasserstein}.

Although GAN-based techniques are exploding, they cannot be directly used in solving ER tasks for three reasons: (1) ER datasets are often highly imbalanced, which aggravates the need of sufficient labeled training data, and may cause the mode collapse problem during the training process; (2) Most of the GAN-based approaches, including the ones designed for semi-supervised learning \cite{springenberg2015unsupervised}, have not considered the case of training with a extremely limited number of real labeled instances; (3) Traditionally, the generator in GANs is designed to generate new instances; however, for ER tasks, classifying all unlabeled instances is the ultimate goal. Until now, there is no existing GAN-based approach that can address all these problems. In this work, we build \textsc{ErGAN} to fill in this gap.

\subsection{Deep Learning for Entity Resolution}
In recent years, motivated by the success of deep learning techniques on tasks in computer vision, natural language processing, etc. \cite{goodfellow2016deep}, several attempts have been made to design deep learning solutions for ER tasks \cite{mudgal2018deep, ebraheem2018distributed}.
Ebraheem et al. proposed DeepER, which uses bi-directional Recurrent Neural Networks (RNNs) with Long Short Term Memory (LSTM) units to learn a distributed representation for each record \cite{ebraheem2018distributed}. Mudgal et al. studied how to use deep learning techniques developed in natural language processing to handle the problems of attribute embedding, attribute summarization and attribute comparison \cite{mudgal2018deep}. A recent work proposed by Nie et al. \cite{nie2019deep} uses an align-compare-aggregate framework for a token level sequence-to-sequence ER which aims to solve the heterogeneous and dirty data problems. 
To deal with limited labels, several approaches take the advantages of transfer learning technique \cite{zhao2019auto,kasai2019low}. However, the well-known limitations of transfer learning are that it needs a pre-trained model before applying to a target task, and a prior assumption on the correlation between the source and target tasks is also required, which restrict its practical applicability for ER problems in real-world applications.

To the best of our knowledge, this work is the first to explore the potentials of using GAN techniques to build powerful ER classifiers without prior knowledge. Our proposed method \textsc{ErGAN} can be synthesized with other deep learning techniques, including word embedding and transfer learning, and used as a key building block of an end-to-end deep learning solution.


\section{Conclusion}
In this paper, we have proposed a novel method, called \textsc{ErGAN}, to solve the ER classification problem with very limited labeled instances. \textsc{ErGAN} incorporates the diversity of instances into sampling, prior to training the models. \textsc{ErGAN} consists of a label generator $G$ to generate pseudo labels for unlabeled instances, and a discriminator $D$ to distinguish instances with pseudo labels from instances with real labels. 
This method can be extended with word embedding for handling attribute values, leading to an enhanced method, called \textsc{ErGAN+WE}.
We have formally proven that even with a limited number of instances with real labels, $D$ and $G$ in \textsc{ErGAN} can converge to the true distribution of instances and their labels. Our experimental results show that the performance of our methods beats all the baselines.

\bibliographystyle{plain}
\bibliography{ER-GAN}

\end{document}